\pdfoutput=1
\documentclass{article} 
\usepackage[final]{colm2026_conference}

\usepackage{microtype}
\usepackage{hyperref}
\usepackage{url}
\usepackage{booktabs}
\usepackage{wrapfig}

\usepackage[utf8]{inputenc} 
\usepackage[T1]{fontenc}    
\usepackage{hyperref}       
\usepackage{url}            
\usepackage{booktabs}       
\usepackage{amsfonts}       
\usepackage{nicefrac}       
\usepackage{microtype}      

\usepackage{algorithm}
\usepackage{algpseudocode}
\usepackage{amsmath,amssymb}
\usepackage{xcolor}         
\usepackage{graphicx}
\usepackage{amsmath, amssymb}
\usepackage{subcaption}
\usepackage{adjustbox}
\usepackage{enumitem}
\usepackage[subtle]{savetrees}
\usepackage{xspace}
\usepackage{multirow}
\usepackage{makecell}
\usepackage[ruled,vlined,noend,algo2e]{algorithm2e}

\usepackage{tabularx, booktabs, array, multirow}

\usepackage{lineno}

\definecolor{darkblue}{rgb}{0, 0, 0.5}
\hypersetup{colorlinks=true, citecolor=darkblue, linkcolor=darkblue, urlcolor=darkblue}

\title{Decoupling Planning and Control for Instructable Agents}

\author{
Zineng Tang \\
\texttt{terran@berkeley.edu} \\
UC Berkeley
\And
Kelsey R.~Allen\thanks{Work done at Google DeepMind.} \\
\texttt{krallen@cs.ubc.ca} \\
UBC
\And
Sjoerd van Steenkiste \\
\texttt{svansteenkiste@google.com} \\
Google DeepMind
\And
Ishita Dasgupta \\
\texttt{idg@google.com} \\
Google DeepMind
\And
Alane Suhr \\
\texttt{suhr@berkeley.edu} \\
UC Berkeley
}

\usepackage{xspace}
\usepackage{amsfonts}
\usepackage{listings}
\usepackage{cite}
\usepackage{multirow}
\usepackage{tabularx}
\usepackage{tcolorbox} 
\usepackage{array} 
\usepackage{caption}

\let\cite\citep

\newcommand{\methodname}{Instruct-to-Act\xspace}

\begin{document}

\ifcolmsubmission
\linenumbers
\fi
\maketitle

\begin{abstract}
Recent work shows that pre-trained, instruction-tuned vision-language models (VLMs) perform well at mapping from instructions and observations to high-level plans, but struggle to realize such plans as reliable low-latency action sequences in unfamiliar environments. At the same time, world-model controllers excel at fast observation-to-action control, but lack open-ended task guidance. In this work, we combine these strengths into a single system, \methodname{}, where we train a world-model controller to act autonomously at high frequency when conditioned on sparse, higher-latency, and high-level text instructions generated by a VLM planner.
To train controllers to be language-instructable, we relabel segments of controller policy rollouts with synthetic instructions and jointly optimize a behavior-cloning objective along with existing reward-maximizing and world-modeling objectives. 
We evaluate our proposed approach across seven embodied environments, including three multi-agent environments where  VLM planners coordinate through language while trained controllers serve as their actuators.
Under matched observation and action spaces, our decoupled approach consistently outperforms controller-only and direct VLM action-generation variants, preserves fast control, and lets us swap in different pretrained VLM planners without fine-tuning, while remaining competitive with strong vision-language-action and multi-agent RL baselines on six of seven tasks.\footnote{Code and a demo for our project is available at \url{https://zinengtang.github.io/speak-to-act/}.}
\end{abstract}


\section{Introduction}
Embodied agents increasingly operate in settings that demand both high-frequency, low-latency continuous control and intermittent, high-level reasoning and planning.
For example, consider a long-horizon, multi-player game like Minecraft, where agents can communicate with one another to achieve shared goals.
There have been two dominant approaches in building such agents.
Learning world models through reinforcement learning (RL) has demonstrated that compact modeling of latent environment dynamics can support sample-efficient, low-level control from pixels~\cite{hafner2020dreamer,hafner2023dreamerv3}. 
In parallel, instruction-following large language and vision-language(-action) (LLM/VLM/VLA) models~\cite{zitkovich2023rt,black2410pi0,kim2024openvla,bjorck2025gr00t} can synthesize useful decompositions, subgoals, and plans from high-level task descriptions and visual context. 
Yet these two capabilities are typically available in isolation. VLM-only agents plan well, but struggle to realize plans as precise action streams under tight real-time budgets. Moreover, because VLM outputs are text tokens rather than time-critical action streams, evaluating them as embodied agents across diverse domains is challenging. Low-level controllers act smoothly and quickly but are difficult to steer with abstract, open-ended instructions.

In this work, we study the problem of grounding language into policy for real-time control coupled with high-level abstraction. 
We propose \methodname{}, a plug-and-play paradigm that decomposes the problem of embodied decision-making into \textit{planning} and \textit{control}.
In this paradigm, a pre-trained VLM \textit{planner} maps environment observations to plans and instructions; these instructions are sent to a \textit{controller}, which processes a stream of observations and produces low-level actions in real time.
The VLM planner plans and issues instructions concurrently with low-level execution, without the need for finetuning. 
Our core contributions center around a framework for efficiently training environment-specific controllers that can be paired with any planner to support real-time planning and control.
Each controller is built as a \textbf{language-aware world model and policy}, where both representation learning (on the world model) and control (actor and critic heads) are conditioned on a latent representation of an instruction sent by a planner.
In contrast with single-task models like Dreamer~\citep{hafner2020dreamer,hafner2023dreamerv3}, where a policy is learned indirectly through rewards, our setting allows latent imagination and value learning to incorporate a wide range of abstract goals specified using a language instruction.
To train controllers to be conditioned on instructions, we perform \textbf{post-hoc instruction supervision}, summarizing replay segments into language instructions with a VLM and using them to perform behavior cloning during training. 
Finally, during inference, we perform \textbf{asynchronous instruction}: the planner continuously reasons and plans in text space while the controller simultaneously executes low-level actions in the environment.

Because planning is decoupled from acting, inference scales well in both performance and efficiency: any number of language planners can each be coupled with identical controller instances, enabling efficient \textbf{multi-agent communication} in real time.
This design also lets us evaluate both (a) how well individual VLMs act as embodied single agents and (b) how well they reason about, coordinate with, and communicate with other agents under real-time constraints.

We organize evaluation into two parts.
First, we verify that without instructions, trained controllers match the performance of standard world-model RL baselines, and that they achieve an average instruction-following accuracy of 92.8\% across instructions generated from a diverse set of VLM planners.
Then, we use the controllers to evaluate VLM planners' ability to (a) act as agents across diverse single-player embodied environments by mapping high-level task descriptions to mid-level instructions, and (b) plan and coordinate with other VLM planners in multi-agent tasks.
Pairing VLM planners with our trained controllers performs significantly better than mapping directly from high-level task to low-level actions, in both task score and inference efficiency, achieving competitive performance in 6 of the 7 environments; overall, decoupling planning and control supports strong task performance, low latency, and convenient modularity for evaluating VLMs as embodied agents in arbitrary heterogeneous environments.

In contrast to prior work (Section~\ref{sec:related}), no VLM is fine-tuned to an environment's low-level action space---only the lightweight controller is environment-specific; the controller, unlike pure world-model RL agents, is explicitly trained to execute open-ended language instructions at control frequency; and the stable language interface supports asynchronous planner swaps and direct multi-agent coordination with a shared controller architecture.

\begin{figure*}[t]
\centering
\includegraphics[width=0.85\textwidth]{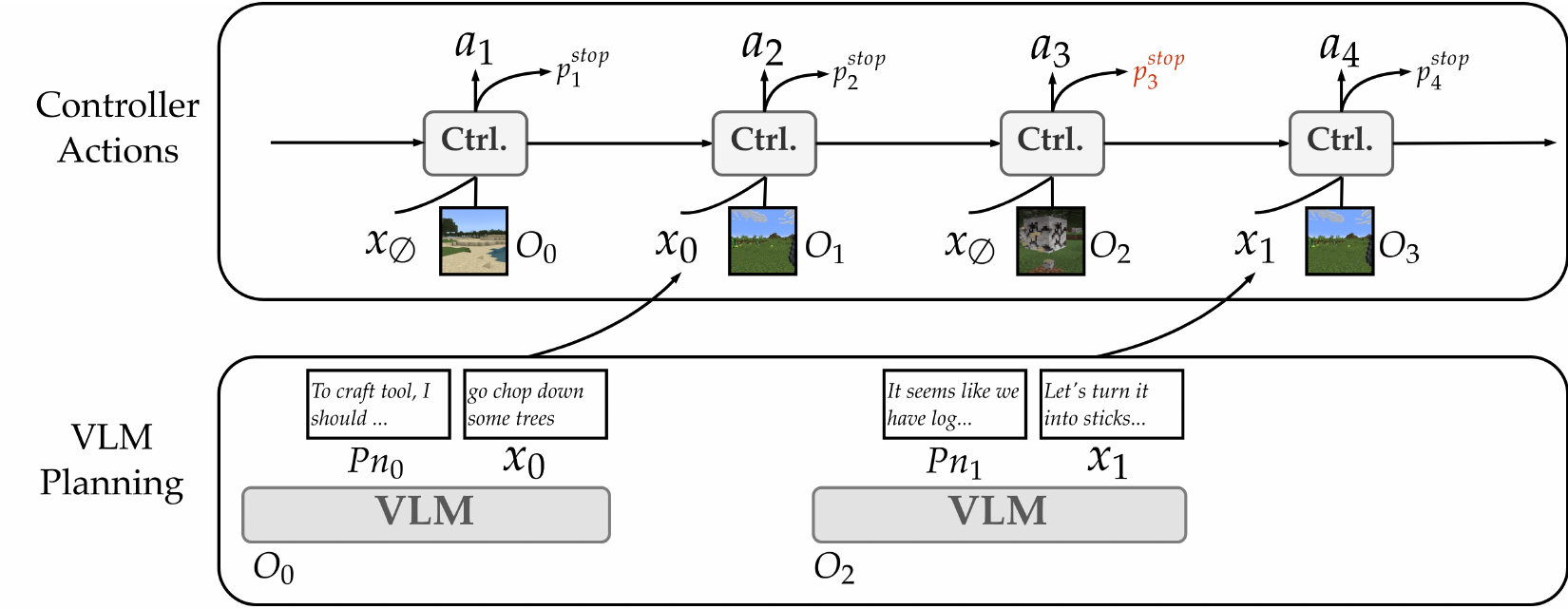}
\caption{Illustration of inference with a VLM planner and trained controller.
}
\label{fig:dynamics_model}
\vspace{-14pt}
\end{figure*}


\section{Overview and Experimental Setup}
\label{sec:setup}
We study embodied agents that map observations to low-level actions, optionally conditioned on natural-language task specifications. Our goal is to adapt existing vision-language models (VLMs) into embodied, language-using agents out-of-the-box, without fine-tuning them to each environment's low-level action space.


At inference time, we decompose decision making into a \textit{planner} and \textit{controller}. The planner maps observations (and optional task context) to a natural-language instruction, and the controller maps the instruction and visual observations to low-level actions. This contrasts with approaches that require end-to-end low-level action generation by fine-tuned VLM/VLA models~\cite{driess2023palm,zitkovich2023rt,black2410pi0,kim2024openvla,bjorck2025gr00t,hafner2020dreamer,hafner2023dreamerv3}, and follows language-conditioned control paradigms in robotics~\citep{Ahn2022DoAI}; it allows us to plug in any instruction-tuned VLM as a planner with a lightweight environment-specific controller.

\paragraph{Task.} We consider partially-observable environments $(\mathcal{S},\mathcal{A},\Omega,O,P,R,\gamma)$ with state space $\mathcal{S}$, action space $\mathcal{A}$, observation space $\Omega$, observation kernel $O: \mathcal{S} \rightarrow \Delta^{\Omega}$, transition kernel $P: \mathcal{S} \times \mathcal{A} \rightarrow \Delta^{\mathcal{S}}$, reward function $R$, and discount factor $\gamma$.
Pretrained VLMs struggle when $\mathcal{A}$ is high-frequency and environment-specific (e.g., repeated camera/control commands in Minecraft or Atari), making direct low-level action sequence emission impractical. In our target domains, the challenge is not only choosing the right subgoal, but sustaining a long, low-level action stream at control rate.

At each environment step, an agent policy $\pi: \Omega \rightarrow \Delta^{\mathcal{A}}$ maps from observations to a distribution over actions.
We study agents which are compositions of independent \textit{planner} and \textit{controller} agents, respectively $\pi_p: \Omega \rightarrow \Delta^\mathcal{X}$ and $\pi_c: \Omega \times \mathcal{X} \rightarrow \Delta^\mathcal{A}$, where $\mathcal{X}$ is the set of all possible natural language utterances. As illustrated in Fig.~\ref{fig:dynamics_model}, this planner–controller split is closely related to hierarchical RL and the options framework, where higher levels issue temporally extended commands~\cite{dayan1992feudal,sutton1999between,bacon2017option,nachum2018data,levy2018hierarchical}.
We sample actions from $\pi$ given an observation $o$ by first sampling an instruction $x \sim \pi_p(\cdot \mid o)$, 
then sampling a sequence of low-level actions $\overline{a}$ from $\pi_c$, where $a_i \sim \pi_c(\cdot \mid o_i, x)$.
We augment the action space with an instruction-completion action; when selected, a new instruction is sampled from $\pi_p$. Controllers need not be conditioned on an instruction; when none is available, the controller can continue autonomously between planner updates. In experiments, $\pi_p$ is an instruction-tuned VLM and $\pi_c$ is a lightweight planner-agnostic environment-specific recurrent state space model (RSSM).

\paragraph{Controller.} The controller $\pi_c$ is an instruction-conditioned RSSM~\cite{hafner2023dreamerv3} with an additional head predicting instruction completion (Sec.~\ref{sec:arch}). It is trained by relabeling segments of its own on-policy rollouts with VLM-generated instructions and adding a behavior-cloning objective to the standard Dreamer world-model, actor, and value losses (Sec.~\ref{sec:generating-trajectories}). No expert demonstrations are used.

\paragraph{Evaluation.} We evaluate on seven environments spanning classic single-agent RL benchmarks~\cite[e.g., Atari, ][]{bellemare2013ale} and multi-agent tasks based on real cooperative video games (e.g., Pico Park); tasks, metrics, and analyses are detailed in Section~\ref{sec:eval}.

\section{Method}

\subsection{Decoupling Planning and Control in Inference}
\label{sec:online_inference}
Our agent policy $\pi$ composes a planner $\pi_p$ and a controller $\pi_c$ (Section~\ref{sec:setup}). In practice, we run inference on $\pi_p$ and $\pi_c$ in independent parallel processes, which exploits $\pi_p$'s capabilities beyond high-level decision-making---reasoning and communication with other agents---while low-level actions execute in the environment, and extends naturally to multi-agent settings by spawning planner and controller threads per agent.

\paragraph{Asynchronous Online Inference.} 
Trading off inference between $\pi_p$ and $\pi_c$, where $\pi_p$ must wait for $\pi_c$ to complete its current instruction before it starts to plan a new one, wastes wall-clock time.
Instead, we allow planners $\pi_p$ to run inference asynchronously in the background, given a live observation stream, continuously updating their reasoning, illustrated by Alg.~\ref{alg:online} (pseudocode in Appendix~\ref{app:algorithm}).
Upon receiving a message that the controller has completed its current instruction, planners are prompted to immediately
emit the next instruction from a draft plan. 
Given a draft plan, emitting an instruction requires generating only a few tokens. 
This simultaneous inference supports both the relatively slow language reasoning of VLMs beneficial to generating optimal instructions, and the fast low-level execution provided by the controller.
In the offline variant (Alg~\ref{alg:offline}), by contrast, the controller waits until the planner finishes producing a new instruction, so throughput directly inherits VLM latency at every instruction boundary.

\paragraph{Multi-agent Inference.} We extend the framework to cooperative tasks with $n$ agents $\left\{\pi^{(i)} = \left(\pi^{(i)}_p, \pi_c\right)\right\}^n_{i=1}$.\footnote{We include more details on the multi-agent setup in Appendix~\ref{app:algorithm}.}
Each agent shares controller parameters, but at inference time, these controllers operate on different observation streams, maintain different latent states, and execute different actions.
Planners need not be the same type of agent.
Unlike classic centralized-training/decentralized-execution~\cite[CTDE, ][]{kraemer2016multi}, we do not train a separate centralized critic or impose structured role labels. Instead, coordination happens purely in language, with planners specifying and negotiating high-level goals.
Agents communicate in discrete rounds aligned with the control step. At each step, the agents can (i) receive messages from the shared chatroom  $\texttt{Chatroom}[i]_t$; (ii) instruct the controller to act; (iii) optionally speak. 
 The chatroom $\texttt{Chatroom}[i]$ concatenates all peer messages at any given step.
All agents share the same planner system prompt, which instructs them to process incoming messages and reply as if in a dialogue.\footnote{Appendix~\ref{app:prompts} contains all system prompts.} Qualitatively, agents use messages to claim subgoals, negotiate roles early in an episode, and report bottlenecks when progress stalls.
Controllers do not process messages sent between planner agents.

\subsection{Controller Architecture}
\label{sec:arch}

Our controller is based on the recurrent state space model~\cite[RSSM; ][]{hafner2019learning,hafner2020dreamer}, which maintains a latent state $s_t$ computed from previous observations $o_{<t}$ and actions $a_{<t}$, with decoder heads estimating the next observation, reward, and action. Following~\citet{hafner2023dreamerv3}, we train the observation decoder with a KL-regularized reconstruction objective and optimize Dreamer-style actor and value heads on imagined rollouts.\footnote{Full equations and losses can be seen in Appendix~\ref{app:worldmodel}.}
We adapt this base architecture to optionally condition on a planner instruction $x_t$: a frozen, pretrained contrastive language encoder maps $x_t$ to $e_t \in \mathbb{R}^{d_\ell}$, which is projected and concatenated with the RSSM features used by the policy and value heads, yielding
$$
\pi(a\mid s_t,e_t)=\mathrm{Softmax}\!\big(\ell(s_t,e_t)/\tau\big)_a, \qquad \ell(s_t,e_t) \in \mathbb{R}^{|\mathcal{A}|}.
$$
When no instruction is available, $e_t$ is a learned null embedding, so the controller can act autonomously.

We additionally augment the output space of the controller to include an indicator of whether an instruction has been completed.
At each step $t$, the actor emits
$
p^{\text{stop}}_t \;=\; \sigma\!\big(g(s_t,e_t)\big)
$
where $g$ is a small head on top of the latent state and $\sigma$ is the logistic function. 

\subsection{Planner}
\label{sec:planner}
Our framework supports any planner $\pi_p: \Omega \rightarrow \Delta^\mathcal{X}$ that maps from observations to instructions, where observations can include visual input and, for multi-task environments, an optional natural-language task specification.
We implement planners using vision-language models (VLMs), prompted to additionally maintain, during inference, a memory bank $Mr$ and a partial plan $Pn$.
Concretely, at inference step $t'$, the VLM maps $(Sp, Mr_{t'-1}, Pn_{t'-1}, O_{<t'}, p^\text{stop}_{t'-1})$---system prompt, memory bank, partial plan, observation history, and the controller's most recent instruction-completion probability---to an updated $(Mr_{t'}, Pn_{t'})$ and, if $p^\text{stop}_{t'-1}$ indicates completion, an instruction $x$ sent to the controller $\pi_c$.
Crucially, as discussed in Sec.~\ref{sec:online_inference}, the VLM steps $t'$ need not be synchronous with the controller's timesteps $t$, supporting online planning of the next instruction while the agent executes its current one; in the synchronous variant, each instruction is generated only after the previous one completes, so $t'$ aligns with instruction boundaries.

In multi-agent settings, planners additionally communicate in one of two modes: \emph{decentralized}, where at each VLM inference step any planner may initiate at most one message to a peer, or \emph{centralized}, where a fixed hub agent sends messages to all other agents (formal definitions in Appendix~\ref{app:training_details}).
Relative to the single-agent setting, the planner input is augmented with the agent's inbox and recent outgoing messages, and the planner output can additionally include a short message for another agent.

\subsection{Training}
\label{sec:generating-trajectories}

We train the controller parameters by alternating between collecting rollouts conditioned on the current policy, and optimizing the policy using both reward, world model, and instruction-following behavior-cloning objectives. 
The losses combine two objectives: (a) instruction following, via behavior cloning on on-policy action sequences post-hoc labeled with VLM-generated instructions; and (b) reward maximization and world modeling, which, following the Dreamer approach~\citep{hafner2023dreamerv3}, align the labeled behavior distribution with what the controller executes at inference time.

\paragraph{Behavior cloning with post-hoc instruction annotation.}
During learning, we use a FIFO replay buffer~\cite{mnih2015human} $\mathcal{D}$ (capacity $|\mathcal{D}|{=}1024$) that stores tuples $\tau_t \equiv (o_t, a_t, r_t, e_t, \mathrm{complete}_t, \mathrm{done}_t)$, where $e_t$ is an optional instruction embedding, $\mathrm{complete}_t$ indicates whether $a_t$ completed the instruction, and $\mathrm{done}_t$ marks episode termination.
To acquire $e_t$, we annotate the buffer online: we sample a set of non-overlapping, variable-length intervals $\mathcal{I}=\{[u_k,v_k]\}_{k=1}^K$ and, for each interval, prompt a VLM summarizer with task-specific prompts, equally spaced frames, and the packed action-sequence text  to produce a high-level instruction $x$. We encode $x$ into $e$ (Section~\ref{sec:arch}), pair all tuples $t \in [u_k, v_k]$ with $e$, and set $\mathrm{complete}_t{=}1$ only at $t{=}v_k$.\footnote{Further segment sampling details and per-timestep bookkeeping can be referred in Appendix~\ref{app:training_details}.}

During training, we treat $a_{u_k:v_k}$ as a label for the instruction $x$ conditioned on observations $o_{u_k:v_k}$, and optimize a behavior cloning loss over labeled intervals:

\begin{small}
\vspace{-0.75em}
$$
\mathcal{L}_{\text{BC}} \;=\; -\,\lambda_{\text{BC}} \sum_{[u_k,v_k] \in \mathcal{I}} \sum_{t=u_k}^{v_k} \log \pi\!\big(a_t \mid s_t, e_t\big).
$$
\vspace{-0.75em}
\end{small}

\noindent To learn to mark instructions as complete, we also apply a binary cross-entropy loss on $p^{stop}_t$ using the label $\mathrm{complete}_t$:

\begin{small}
\vspace{-0.5em}
$$
\mathcal{L}_{\text{stop}} \;=\; -\sum_t \Big(\mathrm{complete}_t\log \left(p^{\text{stop}}_t\right) + \left(1-\mathrm{complete}_t\right)\log \left(1-p^{\text{stop}}_t\right)\Big).
$$
\vspace{-0.5em}
\end{small}

We annotate intervals covering 50\% of the buffer, leaving the remaining unannotated so that the controller also learns to act autonomously from the environment reward $R$. Annotation reuses logged trajectories and runs in parallel with optimization, adding roughly 17\% GPU-hours and a 12\% training slowdown (Section~\ref{sec:experiments}). When no instruction is provided, we use the learned null embedding from Section~\ref{sec:arch}.
We (optionally) support a strict instruction-required mode that architecturally guarantees no actions are executed without a valid, non-completed instruction.

\paragraph{Final Training Objective.} The overall objective augments the base objective with behavior cloning:
$$
\mathcal{L}\;=\;\mathcal{L}_{\text{model}} \;+\; \lambda_V\,\mathcal{L}_{\text{value}} \;+\; \lambda_A\,\mathcal{L}_{\text{actor}} \;+\; \mathcal{L}_{\text{BC}}\;+\; \mathcal{L}_{\text{stop}}
$$
where $\mathcal{L}_\text{model}$, $\mathcal{L}_\text{value}$, and $\mathcal{L}_\text{actor}$ are defined by RSSM training algorithm~\cite{hafner2023dreamerv3}.

\section{Experiments}
\label{sec:experiments}

\subsection{Architecture and Training}

Our controller is a language-conditioned DreamerV3-style RSSM~\citep{hafner2023dreamerv3} with CLIP-base-224~\cite{pmlr-v139-radford21a} and DINOv2-base~\cite{oquab2023dinov2} visual features, and MiniLM-L6-H384-uncased~\citep{wang2020minilm} language embeddings. Main results use the 800M controller variant. During data collection, rollouts are generated with empty language ($e_t=\mathbf{0}$), then post-hoc instruction annotation labels random replay segments (length 1 to 20) while unlabeled gaps carry no behavior-cloning loss. GPT-4o is the annotator and 50\% of replay segments are annotated.\footnote{Hyperparameters, model scales, and compute/runtime overhead are shown in Appendix~\ref{app:hyperparameters}, Appendix~\ref{app:large_scales}, and Appendix~\ref{app:train_compute} respectively.}

\vspace{-8pt}
\subsection{Evaluation}
\label{sec:eval}
Each evaluation consists of 100 episodes (or the task's standard protocol), with no exploration noise ($\varepsilon{=}0$) and temperature $\tau{=}0.8$ for stochastic policies.
We follow each environment's official protocol and metric: Atari~\cite{kaiser2019model} (sticky actions, no-op starts), MineRL ObtainDiamond~\cite{guss2019minerl}, Crafter~\cite{hafner2021benchmarking}, DMLab \texttt{explore\_goal\_locations}~\cite{beattie2016dmlab}, Overcooked~\cite{carroll2019utility} (scripted partners for single-agent tests, cross-play for multi-agent), Pico Park (cooperative puzzle completion with synchronized actions),\footnote{Pico Park (\url{https://picoparkgame.com/en/}) is a 2-8 player 2d cooperative platformer puzzle game.} and MindCraft~\cite{white2025collaborating} (cooking/construction/crafting success).\footnote{Detailed tasks descriptions are in Appendix~\ref{app:task}.  We also include a lightweight human experiment in Appendix~\ref{app:human}, a multi-agent failure mode analysis in Appendix~\ref{app:failure}, an instruction-length robustness study in Appendix~\ref{app:robust}, and a taxonomy of generated instructions in Appendix~\ref{app:taxonomy}.}

\vspace{-8pt}
\subsection{VLMs as Planners}

\paragraph{Baselines Setup.}
\label{sec:baselines}
We group baselines into three categories. Reproduced (DreamerV3, QMIX, MAPPO): trained and evaluated by us under our observation and action spaces, giving frameworks that require extensive language reasoning additional training and inference budget until convergence. Domain-adapted: RT-2, fully fine-tuned on demonstrations from a trained DreamerV3 policy with non-pixel observations tokenized alongside pixels; as this departs from its original real-robot setting, these numbers reflect a domain adaptation. Contextual (Voyager, DEPS, JARVIS-1, LS-Imagine): evaluated with released code on Minecraft Diamond and adapted code on MindCraft, as their prompting infrastructure is Minecraft-specific and new prompts on new environments would introduce confounds. 

\vspace{-5pt}
\paragraph{Results.}
Table~\ref{tab:all_eval_tasks} shows the main results, comparing different VLMs as planners with several baseline agents. 
We train one controller per environment, so training does not require planner-specific fine-tuning or VLM reasoning in the loop. The table includes: (i) our planner+controller agents using four VLM planners; (ii) ablations that remove the controller or language interface; (iii) prior environment-specific baselines; and (iv) recent VLA/MARL baselines when available. 
The main pattern is consistent across tasks: decomposition into \textit{planning} and \textit{control} improves over direct VLM action generation, and adding language guidance improves over the controller-only baseline. Nearly all tested planners significantly outperform the end-to-end GPT-4o agent. Using GPT-4o to generate sparse instructions executed by a trained controller, rather than directly producing low-level actions, improves performance on every task. The strongest planners (GPT-4o and Qwen-VL-2.5-72B) achieve the best overall results, while our method remains competitive with stronger domain-specialized systems such as JARVIS-1 in Minecraft.

\begin{table}[t]
\centering
\adjustbox{width=\textwidth,center}{
\begin{tabular}{>{\raggedright\arraybackslash}p{0.13\textwidth}>{\raggedright\arraybackslash}p{0.35\textwidth}*{7}{c}}
\toprule
 & & \multicolumn{4}{c}{\textbf{Single-Agent Tasks}} & \multicolumn{3}{c}{\textbf{Multi-agent Tasks}} \\
\cmidrule(lr){3-6} \cmidrule(lr){7-9}
  & & \textbf{Atari} & \textbf{MC-Dia} & \textbf{Craft} & \textbf{DML} & \textbf{OCook} & \textbf{Pico} & \textbf{MindC} \\
\midrule
\multirow{4}{*}{\makecell[l]{\textbf{Ours}\\(planner\\ swap)}}
& Gemma-3-27B~\cite{gemma_2025} & 880 & 9.7 & 13.8 & 71 & 187.4 & 68.9 & 53.0 \\
& llava-v1.6-34b~\cite{liu2024llavanext} & 862 & 9.9 & 12.8 & 74 & 187.2 & 63.5 & 51.6 \\
& Qwen-VL-2.5-72B~\cite{bai2025qwen25vl} & 878 & 11.1 & 13.4 & 77 & 192.3 & 68.4 & 58.5 \\
& GPT-4o~\cite{hurst2024gpt} & 891 & 11.7 & 14.1 & 76 & 193.2 & 70.1 & 70.2 \\
\midrule
\multirow{3}{*}{\textbf{Ablation}}
& Qwen-VL-2.5-72B (Direct VLM Finetune) & 581 & 10.1 & 7.6 & 45 & 150.1 & 30.6 & 38.2 \\
& GPT-4o (w/o controller) & 670 & 10.4 & 8.7 & 56 & 180.4 & 50.3 & 50.2\\
& Controller-Only & 809 & 8.2 & 12.6 & 67 & 170.2 & 30.7 & 40.0 \\
\midrule
\multirow{3}{*}{\makecell[l]{\textbf{Reproduced}\\\textbf{Baselines}\\}}
& Dreamerv3~\cite{hafner2023dreamerv3} & 811 & 8.6 & 10.5 & 65 & - & - & - \\
& MAPPO~\cite{yu2022surprising} & - & - & - & - & 182.5 & 50.8 & 44.9 \\
& QMIX~\cite{rashid2020qmix} & - & - & - & - & 187.2 & 58.5 & 53.2 \\
\midrule
\makecell[l]{\textbf{Domain-}\\\textbf{Adapted}\\(\S\ref{sec:baselines})}
& RT-2~\cite{zitkovich2023rt} & 457 & 6.2 & 4.7 & 36 & 124.7 & 34.0 & 33.1 \\
\midrule
\multirow{5}{*}{\makecell[l]{\textbf{Contextual}\\\textbf{Comparisons}\\}}
& Voyager~\cite{wang2023voyager} & - & 11.8 & - & - & - & - & - \\
& DEPS~\cite{wang2023deps} & - & 9.4 & - & - & - & - & 45.6 \\
& LS-Imagine~\cite{li2025lsimagine} & - & 9.6 & - & - & - & - & 50.1 \\
& JARVIS-1~\cite{wang2024jarvis1} & - & 12.3 & - & - & - & - & 54.1 \\
& Mindcraft~\cite{white2025collaborating} & - & - & - & - & - & - & 49.0 \\
\bottomrule
\end{tabular}}
\caption{Main results. Each cell reports the task's official metric where available. Baselines are grouped by comparability into \textbf{reproduced}, \textbf{domain-adapted}, and \textbf{contextual} categories, detailed in Section~\ref{sec:baselines}; our controlled claim rests on the matched comparison between our planner-controller variants and the ablation rows. Abbr: MC-Dia (Minecraft Diamond), Craft (Crafter), DML (DMLab), OCook (Overcooked), Pico (Pico Park), and MindC (MindCraft).}
\label{tab:all_eval_tasks}
\vspace{-10px}
\end{table}

\subsection{Scalability and Efficiency}
\label{sec:scaling}
\begin{figure*}[t]
  \centering
  \begin{subfigure}{0.499\textwidth}
    \centering
    \includegraphics[width=\linewidth]{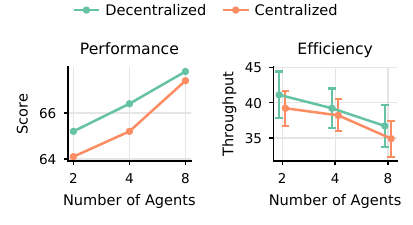}
    \vspace{-10pt}
    \caption{Agent scaling (Pico Park)}
    \label{fig:agent_scaling}
  \end{subfigure}\hfill
  \begin{subfigure}{0.499\textwidth}
    \centering
    \includegraphics[width=\linewidth]{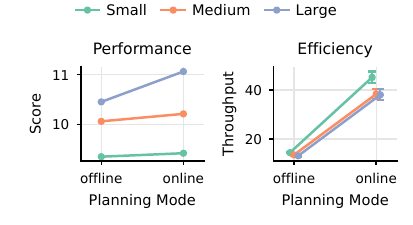}
    \vspace{-10pt}
    \caption{Planning modes (Minecraft Diamond)}
    \label{fig:planning_modes}
  \end{subfigure}
  \caption{Per-setting \emph{performance} (left) and \emph{efficiency} (right) results for (a) agent scaling and (b) planning modes. Model-size and reasoning scaling curves are shown in Figure~\ref{fig:perf-vs-eff-pairs} (Appendix~\ref{app:scaling_efficiency_plots}).}
  \label{fig:perf-vs-eff-main}
  \vspace{-10pt}
\end{figure*}

We evaluate the framework's performance and efficiency (throughput as environment steps per second) under multiple conditions.\footnote{Additional conditions are described in Appendix~\ref{app:scaling_efficiency_plots} and Figure~\ref{fig:perf-vs-eff-pairs}.}
In Figure~\ref{fig:perf-vs-eff-main}a, we compare decentralized with centralized control in the multi-agent Pico Park environment. 
Each agent is implemented as a Qwen-VL-2.5-72B planner paired with an 800M-parameter controller.
We measure task performance and throughput as we vary the number of agents from 2 to 8. 
We find that decentralized communication results in improved task performance while also maintaining a throughput equivalent to centralized control.
This is consistent with the motivation for centralized-training/decentralized-execution schemes in MARL, where routing all decisions through a central entity scales poorly with the number of agents~\citep{kraemer2016multi,rashid2020qmix}; in our framework the analogous bottleneck appears at the hub planner in centralized mode, whose failure modes we analyze in Appendix~\ref{app:failure}. Unlike CTDE, however, our agents coordinate through language at the planner level rather than through a centralized critic.

In Figure~\ref{fig:perf-vs-eff-main}b, we compare online with offline planning modes on Minecraft Diamond. 
Here we use a Qwen-VL-2.5~\citep{bai2025qwen25vl} planner with three sizes: small (50M controller with 2B planner), medium (200M controller with 7B planner), and large (800M controller with 72B planner). 
As model sizes increase, online planning offers increased task performance compared to offline planning, while also remaining an order of magnitude faster in terms of throughput.



\subsection{Generalization}
\label{sec:generalization}
We also study how far the framework generalizes beyond the exact configuration it was trained with: does a controller trained on one planner's annotations transfer to other planners and to different instruction-timing policies, do the gains depend on our specific controller architecture, and how reliably does the controller execute held-out planner instructions?
Across these studies,\footnote{Full tables are included in Appendix~\ref{app:large_scales}.} planners that decide instruction timing outperform fixed cadence on Minecraft Diamond (e.g., GPT-4o: $10.75$ for VLM-decided proactiveness vs. $9.60$ fixed cadence), and train\,$\times$\,eval cross-planner transfer remains stable (Table~\ref{tab:train_eval_matrix}). The framework is also not tied to RSSM specifically: a transformer world model is comparable to ours ($11.2$ vs. $11.0$), while weaker controller variants degrade performance (RNN policy: $9.7$; value-only world model: $10.2$). Controller reliability remains high across domains and planners, with instruction-following accuracy between $172/200$ and $194/200$ ($86\%$ to $97\%$).


\begin{table}[t]
\centering
\begin{minipage}[t]{0.53\textwidth}
\centering
\adjustbox{width=\linewidth,valign=t}{
\begin{tabular}{lc}
\toprule
\textbf{Train-time instruction type} & \textbf{Diamond} $\uparrow$\\
\midrule
VLM-generated & 11.1 \\
Template-based instructions & 9.9 \\
Clustered action labels & 8.9 \\
Random strings & 8.3 \\
No instructions & 8.2 \\
Environment-derived labels & 11.4 \\
\bottomrule
\end{tabular}}
\vspace{-5px}
\caption{Impact of post-hoc instruction source on Minecraft Diamond, using a Qwen-VL-2.5-72B planner and an 800M-parameter controller.}
\label{tab:annotation_ablation}
\end{minipage}
\hfill
\begin{minipage}[t]{0.4\textwidth}
\centering
\adjustbox{width=\linewidth,valign=t}{
\begin{tabular}{@{\extracolsep{\fill}}lccc}
\toprule
 & \multicolumn{3}{c}{\textbf{Eval Planner} $\uparrow$} \\
\cmidrule(lr){2-4}
\textbf{Annotator} & \textbf{Gemma} & \textbf{Qwen} & \textbf{GPT} \\
\midrule
Gemma & 9.6 & 9.2 & 10.2\\
Qwen  & 9.7 & 9.9 & 9.9 \\
GPT   & 9.8 & 9.6 & 10.0 \\
\bottomrule
\end{tabular}}
\vspace{-5px}
\caption{Cross-planner train\,$\times$\,eval matrix (Minecraft Diamond). Each cell is score of a controller trained on instructions annotated by the row planner and evaluated by the column planner.}
\label{tab:train_eval_matrix}
\end{minipage}
\vspace{-10pt}
\end{table}
 
\paragraph{Annotation quality.} 
Table~\ref{tab:annotation_ablation} shows the influence of annotation quality during controller training on controller quality. 
We compare VLM-generated instructions to template-generated and cluster-based instructions, as well as environment-derived symbolic labels. 
Environment-derived symbolic labels slightly outperform VLM-generated instructions (11.4 vs.\ 11.1), but these require access to environment metadata, whereas VLM-based annotation needs only raw observations and thus generalizes to domains without such metadata. 
The results show that controllers benefit from semantic content even if instructions are of lower quality; using random strings as instructions, or no instructions at all, significantly underperform template- or cluster-based instructions.

\paragraph{Cross-planner generalization.}
Train\,$\times$\,eval transfer remains stable across all planner pairs in Table~\ref{tab:train_eval_matrix} ($9.2$ to $10.2$), indicating the controller does not overfit to the phrasing of its annotating planner.

\paragraph{Controller cost and scope.}
Our main results use one 800M-parameter controller trained per environment, a one-time cost of $\sim$23 GPU-hours on 4 RTX A6000 GPUs ($\sim$5.5M environment steps; full compute and annotation overhead in App.~\ref{app:train_compute}). In early experiments, a single controller trained jointly on all environments lost 1.12 points on Minecraft Diamond ($\approx$10\% relative), so lightweight domain-general instruction-conditioned controllers remain future work.

\section{Related Work}
\label{sec:related}
Our controller builds on RSSMs and imagination-based control: PlaNet introduced stochastic--deterministic latent transitions for planning from pixels \citep{hafner2019learning}, and Dreamer learns policies by backpropagating value gradients through imagined latent trajectories \citep{hafner2020dreamer,hafner2023dreamerv3}; we instantiate this line of work with language conditioning and post-hoc instruction supervision.

Decoupling slow, deliberate reasoning from fast, reactive control is a dominant paradigm across embodied domains, including autonomous driving \citep{tian2024drivevlm,qian2024fasionad,liu2025vlmudmc}, robot manipulation \citep{zhang2024hirt,shi2025hirobot,bjorck2025gr00t,figure2024helix,song2025hume,chen2025fastinslow}, and vision-and-language navigation \citep{wei2025dualvln}; HiRT \citep{zhang2024hirt} in particular runs an asynchronous planner-controller scheme communicating through latent representations. We do not claim the slow/fast decomposition as novel; we differ in that (i) the planner-controller interface is natural language rather than latents or waypoints, making the planner plug-and-play and enabling language-based multi-agent coordination; (ii) the fast system is a world-model controller rather than a domain-specific stack or waypoint follower, so it can generalize across heterogeneous domains; and (iii) our controller requires no expert data---all behavior-cloning supervision comes from post-hoc VLM relabeling of its own on-policy rollouts.

Language-conditioned robot-control systems motivate our planner-controller decomposition: SayCan, Code as Policies, and Inner Monologue compose language-model reasoning with low-level skills \citep{Ahn2022DoAI,liang2023code,huang2022inner}, while end-to-end VLA approaches such as RT-1, RT-2, RT-H, OpenVLA, $\pi_0$, GR00T, and Octo map observation-language inputs directly to actions \citep{brohan2022rt,zitkovich2023rt,belkhale2024rth,kim2024openvla,black2410pi0,bjorck2025gr00t,team2024octo}. Unlike monolithic VLA adaptation, we train only the smaller environment-specific controller and keep the planner plug-and-play, enabling planner swaps without running full VLM inference at control frequency.
For multi-agent control, we replicate one shared-parameter controller across agents with separate recurrent states, similar to prior MARL work \citep{terry2021revisitingps,chu2017psddpg,rashid2020qmix}; unlike CTDE-style methods with centralized critics or value factorization, our low-level controllers stay decentralized and coordination happens in language at the planner level.

Embodied planner--controller agents in open-ended domains (Voyager, DEPS, LS-Imagine, JARVIS-1 \citep{wang2023voyager,wang2023deps,li2025lsimagine,wang2024jarvis1}) and iterative-reasoning LM agents (ReAct, Reflexion~\cite{yao2022react,shinn2023reflexion}) demonstrate the promise of language-guided planning, but are typically tightly coupled to particular domains, planner backbones, or action-generation pipelines; we instead emphasize a reusable language interface between planner and controller, asynchronous inference, and evaluation across both single-agent and multi-agent domains.

Language is not the only possible conditioning interface: latent goal or skill abstractions such as LISA \citep{garg2022lisa} provide an alternative, but language lets humans and off-the-shelf VLMs act as planners without retraining the controller-side interface, which is central to our plug-and-play evaluation setup.

\section{Conclusion}
We introduced \methodname{}, a framework that combines the low latency of world-model controllers with the high-level reasoning ability of VLMs. Across diverse embodied tasks, the resulting planner+controller agents achieve strong performance while preserving high control-rate throughput, and the same interface extends naturally to multi-agent coordination without requiring specialized MARL training.
Important next steps include studying human planners directly in the loop, extending the framework to asymmetric multi-agent roles and multi-task controller training, and improving planner--controller alignment through prompting or lightweight planner adaptation. More broadly, the framework suggests a practical path for deploying general-purpose language models in domains where abstract guidance must be translated into long, low-level action sequences.

\bibliography{colm2026_conference}
\bibliographystyle{colm2026_conference}

\appendix
\newpage
\section{Model Details}
\subsection{World-model and control details}
\label{app:worldmodel}

\paragraph{RSSM transition and inference.}
Conditioning on language, the deterministic transition and stochastic prior/posterior are
$$
\begin{aligned}
h_t &= f_\theta\!\left(h_{t-1}, z_{t-1}, a_{t-1}, e_{t-1}\right), \\
p_\theta(z_t \mid h_t) &= \mathcal{N}\!\big(\mu_\theta(h_t), \mathrm{diag}\,\sigma^2_\theta(h_t)\big),\\
q_\phi(z_t \mid h_t, o_t, e_t) &= \mathcal{N}\!\big(\mu_\phi(h_t, o_t, e_t), \mathrm{diag}\,\sigma^2_\phi(h_t, o_t, e_t)\big).
\end{aligned}
$$
We denote the model state as $s_t = (h_t, z_t)$ with $z_t \sim q_\phi(\cdot)$ during training and $z_t \sim p_\theta(\cdot)$ during imagination.

\paragraph{Decoders and predictors.}
From $s_t$ we decode/predict:
$$
\begin{aligned}
p_\theta(o_t \mid s_t) \quad &\text{(observation decoder)},\\
p_\theta(r_t \mid s_t) \quad &\text{(reward model)},\\
p_\theta(c_t \mid s_t) \quad &\text{(continuation/discount model with } c_t \in (0,1)\text{)},\\
p_\theta(x_{t+1} \mid s_t) \quad &\text{(next-instruction predictor; optional teacher forcing on } x_t\text{)}.
\end{aligned}
$$
The instruction head is trained only when $m_{t+1}=1$; otherwise its loss is masked out.

\paragraph{Model learning objective.}
Over a sequence $t=1{:}T$, the model loss combines reconstruction/prediction terms with a KL regularizer:
$$
\begin{aligned}
\mathcal{L}_{\text{model}}
&= \sum_{t=1}^T \mathbb{E}_{q_\phi}\Big[
-\log p_\theta(o_t \mid s_t)
-\lambda_r \log p_\theta(r_t \mid s_t)
-\lambda_c \log p_\theta(c_t \mid s_t)
-\lambda_x \, m_{t+1}\, \log p_\theta(x_{t+1} \mid s_t)
\Big] \\
&\quad + \beta \sum_{t=1}^T \mathrm{KL}\!\left(q_\phi(z_t \mid h_t, o_t, e_t)\,\big\|\, p_\theta(z_t \mid h_t)\right),
\end{aligned}
$$
with weights $\lambda_r,\lambda_c,\lambda_x \ge 0$ and KL scale $\beta$.
\subsection{Dreamer-style Control in Latent Space}
Given a learned world model, we learn a policy and value on latent states conditioned on language:
$$
\pi_c(a_t \mid s_t, e_t), \qquad V_\psi(s_t, e_t).
$$

\paragraph{Imagined rollouts.}
Starting from posterior states $s_t$ on real trajectories, we `imagine' $H$-step futures using the prior dynamics and current policy:
$$
\tilde{s}_{t+1} \sim p_\theta(\,\cdot \mid \tilde{h}_{t+1}\,),\;
\tilde{h}_{t+1} = f_\theta(\tilde{h}_{t}, \tilde{z}_{t}, \tilde{a}_{t}, \tilde{e}_{t}),\;
\tilde{a}_{t} \sim \pi_\eta(\cdot \mid \tilde{s}_{t}, \tilde{e}_{t}),\;
\tilde{r}_{t} \sim p_\theta(\cdot \mid \tilde{s}_{t}),\;
\tilde{c}_{t} \sim p_\theta(\cdot \mid \tilde{s}_{t}),
$$
where $\tilde{e}_{t}$ is the instruction embedding available to the agent during imagination (e.g., last known instruction embedding, or an imagined instruction from $p_\theta(x_{t+1}\!\mid s_t)$).

\paragraph{Value targets via $\lambda$-returns with continuation.}
Define the continuation as the learned discount $\gamma \tilde{c}_{\tau} \in [0,1]$. The $\lambda$-return is
$$
\hat{G}_\tau^\lambda
= \tilde{r}_\tau + \gamma \tilde{c}_\tau \Big((1-\lambda)\, V_\psi(\tilde{s}_{\tau+1}, \tilde{e}_{\tau+1}) + \lambda\, \hat{G}_{\tau+1}^\lambda\Big),
\quad
\hat{G}_{t+H}^\lambda \!=\! V_\psi(\tilde{s}_{t+H}, \tilde{e}_{t+H}).
$$

\paragraph{Actor and value losses.}
We optimize the value to regress to the return and the actor to maximize it through imagined trajectories:
$$
\begin{aligned}
\mathcal{L}_{\text{value}}
&= \sum_{\tau=t}^{t+H-1} \big\| V_\psi(\tilde{s}_\tau, \tilde{e}_\tau) - \mathrm{stopgrad}(\hat{G}_\tau^\lambda)\big\|_2^2,\\[2pt]
\mathcal{L}_{\text{actor}}
&= - \sum_{\tau=t}^{t+H-1} \mathbb{E}_{\tilde{a}_\tau \sim \pi_\eta}\Big[\mathrm{stopgrad}(\hat{G}_\tau^\lambda)\Big]
- \alpha \sum_{\tau=t}^{t+H-1} \mathcal{H}\!\left(\pi_\eta(\cdot \mid \tilde{s}_\tau, \tilde{e}_\tau)\right),
\end{aligned}
$$
with entropy scale $\alpha \ge 0$. Gradients backpropagate through the imagined dynamics (world model) as in Dreamer.

\subsection{Model Hyperparameters}
\label{app:hyperparameters}
We use MiniLM-L6-H384-uncased~\citep{wang2020minilm} as the language encoder.
For small controller as an example, we instantiate a language-conditioned DreamerV3-style~\citep{hafner2023dreamerv3} agent with a CNN encoder (4 convolutional blocks; channels $[32,64,128,256]$, stride~2, \textsc{silu}), an RSSM with a deterministic GRU core of size $512$ and a diagonal-Gaussian stochastic latent $z_t\in\mathbb{R}^{32}$, and decoders for observation/reward/continuation as MLPs (two layers, $512$ units, \textsc{silu}). The actor and value heads operate in latent space and use three-layer MLPs with $512$ units (\textsc{silu}). 
The policy head produces action logits for discrete control or mean/scale for continuous control, depending on the environment action space. We follow dreamer to use KL balancing with scale $\beta{=}1.0$ and free-nats $1.5$, reconstruction/prediction weights $(\lambda_r,\lambda_c,\lambda_x){=}(1,1,1)$, imagined horizon $H{=}15$ (as above), and continuation-based $\lambda$-returns. The language pathway pools a frozen text encoder into an embedding $e_t$ that conditions both the world model and control; when absent, a learned null embedding is used. Optimizer and update schedule follow the hyperparameters above (Adam, $3\mathrm{e}{-4}$, clip~$40$, 1:1 world-model/control updates).

Table.~\ref{tab:model_scales} show controller scales from 50M, 200M, to 800M. These configs are adapted from Dreamerv3~\cite{hafner2023dreamerv3} and re-used the 50M and 200M configuration. 

\subsection{Training Compute and Annotation Overhead}
\label{app:train_compute}
Unless otherwise stated, GPT-4o is used as the replay annotator and 50\% of replay segments are instruction-annotated. Training uses 4$\times$ NVIDIA RTX A6000 GPUs and an Intel Xeon Gold 6338 CPU. Across all tasks, the maximum training time for one environment is approximately 36 wall-clock hours (23 hours on average). Relative to controller-only training, post-hoc annotation adds roughly 17\% GPU hours and a 12\% training slowdown.

\subsection{Additional Results}
\label{app:large_scales}

\begin{table}[t]
\centering
\caption{Model-scale configurations.}
\label{tab:model_scales}
\begin{tabular}{lrrrrrr}
\toprule
Scale (tag) & \texttt{rssm.deter} & \texttt{rssm.hidden} & \texttt{rssm.classes} & \texttt{depth} & \texttt{units} \\
\midrule
50M (\texttt{size50m})   & 4096  & 512   & 32  & 32  & 512  \\
200M (\texttt{size200m}) & 8192  & 1024  & 64  & 64  & 1024 \\
800M (\texttt{size800m}) & 24576 & 3072  & 192 & 192 & 3072 \\
\bottomrule
\end{tabular}
\end{table}

\begin{table}[t]
\centering\footnotesize
\begin{tabular}{lc}
\toprule
\textbf{Method} & \textbf{Throughput} \\
\midrule
RT-2~\cite{zitkovich2023rt} & 16.2 \\
DEPS~\cite{wang2023deps} & 20.5 \\
LS-Imagine~\cite{li2025lsimagine} & 23.7 \\
JARVIS-1~\cite{wang2024jarvis1} & 26.0 \\
\methodname{} (Qwen-2.5-VL-72B) & 31.7 \\
\bottomrule
\end{tabular}
\caption{Inference throughput (environment steps/s) for representative VLA baselines and our framework with a Qwen-based planner.}
\label{tab:throughput_baselines}
\end{table}

\paragraph{Efficiency comparison with other works:} Table~\ref{tab:throughput_baselines} complements the throughput plots with a direct comparison against representative end-to-end or tightly coupled baselines. Because our controller executes at every step while the planner is queried only sparsely, \methodname{} achieves the best throughput among these methods.


\begin{table}[t]
\centering
\adjustbox{width=0.6\textwidth,center}{
\begin{tabular}{@{\extracolsep{\fill}}lccc}
\toprule
 & \multicolumn{3}{c}{\textbf{VLM Planner}} \\
\cmidrule(lr){2-4}
\textbf{Instruction Mode} & \textbf{Gemma} & \textbf{Qwen} & \textbf{GPT} \\
\midrule
Fixed Cadence & 8.86 & 9.32 & 9.60 \\
Fully controlled by VLM & 9.89 & 10.12 & 10.52 \\
Proactiveness decided by VLM & 10.12 & 10.23 & 10.75 \\
\bottomrule
\end{tabular}}
\vspace{-5px}
\caption{Single-agent tests under different instruction modes. (Minecraft Diamond)}
\label{tab:minecraft_instruction_modes}
\end{table}

\paragraph{Instruction modes (Minecraft):} We contrast three instruction regimes (Table~\ref{tab:minecraft_instruction_modes}): (1) \emph{Fixed cadence}: planners issue instructions every $K{=}16$ steps; (2) \emph{VLM-decided proactiveness} (the VLM emits instructions only when its uncertainty exceeds a threshold, with a minimum gap of $32$ steps); (3) \emph{Fully controlled by VLM}. We report achievement scores for three planners.


\begin{table}[t]
\centering
\caption{Controller architecture ablation on Minecraft Diamond. The gains are not specific to an RSSM controller, though world-model controllers perform best.}
\label{tab:controller_ablation}
\begin{tabular}{lc}
\toprule
\textbf{Controller architecture} & \textbf{Diamond} \\
\midrule
Dreamer-V3-style RSSM (ours) & 11.0 \\
Transformer world model~\cite{chen2022transdreamer} & 11.2 \\
RNN policy (model-free) & 9.7 \\
Value-only world model & 10.2 \\
\bottomrule
\end{tabular}
\end{table}

\paragraph{Controller architecture:} Table~\ref{tab:controller_ablation} shows that the broader framework is not tied to Dreamer specifically. A transformer world-model controller yields similar trends, whereas removing the world-model structure degrades performance.

\begin{table}[t]
\centering
\caption{Instruction-following accuracy measured on 200 held-out planner instructions per domain.}
\label{tab:instruction_success}
\begin{tabular}{lccc}
\toprule
\textbf{Task} & \textbf{Gemma} & \textbf{Qwen} & \textbf{GPT-4o} \\
\midrule
Minecraft Diamond & 193/200 & 190/200 & 194/200 \\
Pico Park & 186/200 & 180/200 & 189/200 \\
Atari & 181/200 & 172/200 & 185/200 \\
\bottomrule
\end{tabular}
\end{table}

\paragraph{Human Evaluated Instruction Following Success:} We perform human eval in Table~\ref{tab:instruction_success} that quantifies controller reliability under planner instructions. Across domains and planners, instruction-following accuracy stays around $86$--$97\%$. The most common failures arise from vague or underspecified instructions such as ``explore a bit more'' or ``try something different'', which can lead to locally suboptimal behavior until a clearer instruction arrives.

\subsection{Scalability and Efficiency Curves}
\label{app:scaling_efficiency_plots}
\begin{figure*}[t]
  \centering

  \begin{subfigure}{0.499\textwidth}
    \centering
    \begin{minipage}{0.498\linewidth}
      \includegraphics[width=\linewidth]{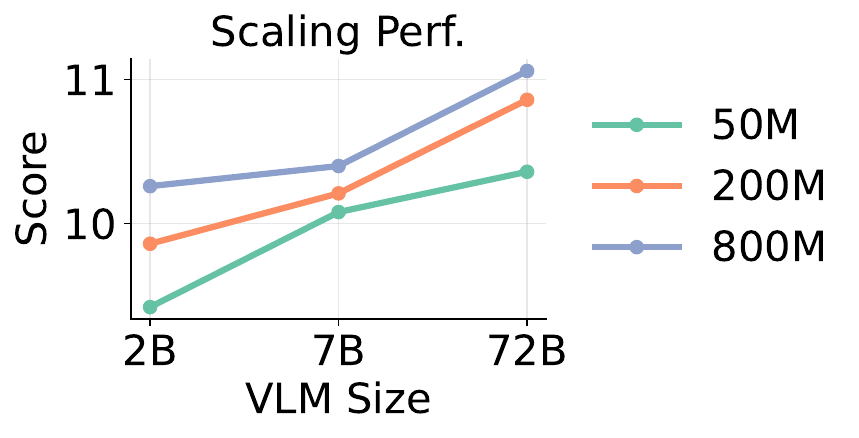}
    \end{minipage}\hfill
    \begin{minipage}{0.498\linewidth}
      \includegraphics[width=\linewidth]{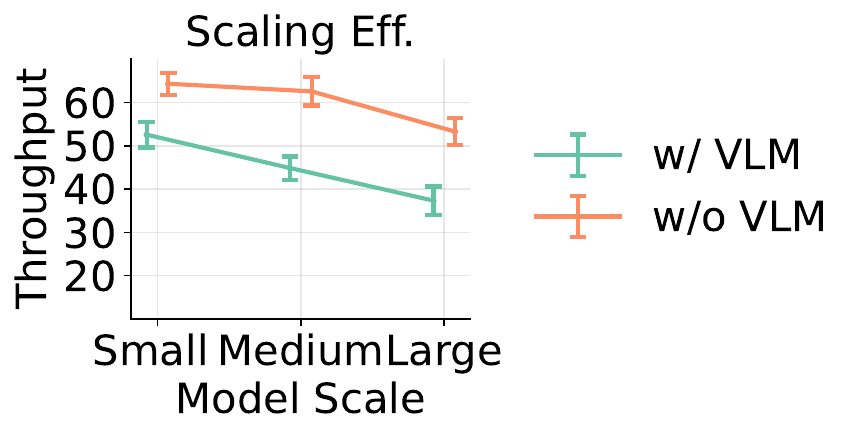}
    \end{minipage}
    \caption{Model size scaling (Minecraft Diamond)}
  \end{subfigure}\hfill
  \begin{subfigure}{0.499\textwidth}
    \centering
    \begin{minipage}{0.498\linewidth}
      \includegraphics[width=\linewidth]{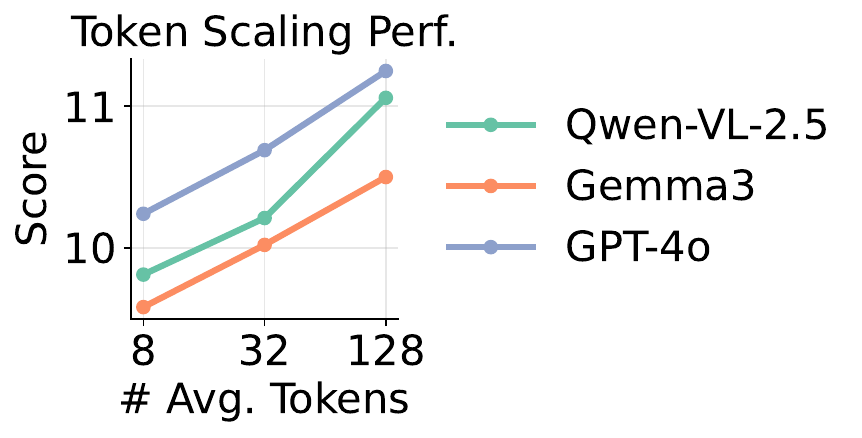}
    \end{minipage}\hfill
    \begin{minipage}{0.498\linewidth}
      \includegraphics[width=\linewidth]{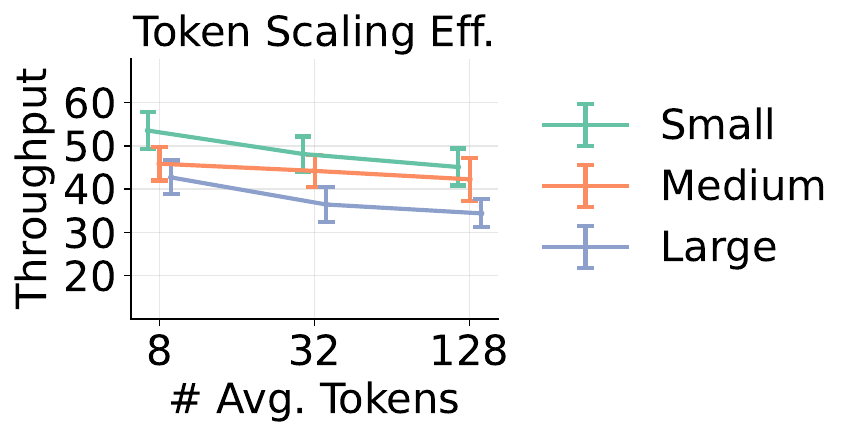}
    \end{minipage}
    \caption{Reasoning scaling (Minecraft Diamond)}
  \end{subfigure}
  \caption{Per-setting \emph{performance} (left) and \emph{efficiency} (right) results for the model-size and reasoning scaling analyses in Sec.~\ref{sec:scaling}. Agent scaling and planning modes are shown in Figure~\ref{fig:perf-vs-eff-main} in the main paper. Model scales for controller and VLM are small (50M, 2B), medium (200M, 7B), and large (800M, 72B).}
  \label{fig:perf-vs-eff-pairs}
\end{figure*}
Figure~\ref{fig:perf-vs-eff-main} (main paper) and Figure~\ref{fig:perf-vs-eff-pairs} report paired curves for each scaling axis, with performance on the left and throughput (environment steps/s) on the right.
Across all four settings, we observe a favorable frontier: configurations that improve task performance usually retain most of the efficiency gains from decoupled control rather than collapsing to the latency profile of end-to-end VLM action generation.
Regarding relative planner performance, Figure~\ref{fig:perf-vs-eff-pairs}(a) provides a controlled scaling study in which every planner-controller combination shares matched settings; there, performance on Minecraft Diamond improves monotonically as either planner or controller size grows. Deviations from a pure parameter-count ordering in Table~\ref{tab:all_eval_tasks} reflect differences in planner pre-training rather than the framework. We also note that Voyager is engineered specifically for Minecraft Diamond, which explains its strong performance on that single task, whereas our framework targets robustness across heterogeneous environments with a single recipe.

\begin{table}[t]
\centering
\small
\begin{tabular}{lcc|cc}
\toprule
 & \multicolumn{2}{c|}{Minecraft Diamond} & \multicolumn{2}{c}{Overcooked} \\
 & Online & Offline & Online & Offline \\
\midrule
VLM latency mean (ms) & 1037.4 & 1221.9 & 1106.3 & 1010.4 \\
VLM latency p95 (ms) & 1701.1 & 2035.7 & 1872.5 & 1523.7 \\
Staleness mean (steps) & 51.2 & 0.0 & 165.1 & 0.0 \\
Blocking steps mean & 0.0 & 35.4 & 0.0 & 0.0 \\
Throughput (env steps/sec) & 37.5 & 13.4 & 169.9 & 92.7 \\
\bottomrule
\end{tabular}
\caption{Latency, instruction staleness, and throughput for online (asynchronous) versus offline (blocking) planning. Overcooked staleness and latency are averaged over both agents.}
\label{tab:async}
\end{table}

\paragraph{Asynchronous planning efficiency.}
Table~\ref{tab:async} quantifies the benefit of asynchrony beyond the two-bar comparison in Figure~2(b). Online planning achieves 2.8$\times$ higher throughput on Minecraft Diamond and 1.8$\times$ on Overcooked (whose shorter episode horizon limits the gain). Asynchrony introduces instruction staleness of roughly 51 steps (Minecraft) and 165 steps (Overcooked) between a stop signal and the arriving instruction, but because the controller continues executing the previous instruction during this window, staleness affects only the next high-level instruction and never low-level control.

\subsection{Multi-agent Failure Mode}
\label{app:failure}
Centralized settings tend to bottleneck on the hub planner, while decentralized settings most often fail through conflicting or duplicated subgoal allocation. To quantify this, we manually examined 100 failure cases in Overcooked, defining controller-level failures as instruction-following errors or stale loops, and planner-level failures as duplicated task assignment, unmet preconditions (e.g., serving a soup that is not yet made), or pipeline stalls. Planner-level coordination failure dominates: 96 of 100 failure windows show near-duplicate sub-task assignments across agents (instruction-embedding cosine similarity above 0.7), while only 3 show controller instruction-following failures.

\subsection{Taxonomy of Generated Instructions}
\label{app:taxonomy}
Across our Minecraft experiments we collected 47{,}547 instruction instances (3{,}477 unique). The set is dominated by local action descriptions (82.8\%, $n{=}39{,}363$), short imperatives describing a single immediate action such as ``Mine stone blocks with a wooden pickaxe'' or ``Swim across the river towards the trees.'' The remainder (17.2\%, $n{=}8{,}184$) are high-level subgoals that chain a primary action to a downstream task, e.g., ``Mine stone to craft a stone pickaxe.'' We observed no retrospective summaries of behavior.

\subsection{Instruction-length Robustness}
\label{app:robust}
We generated 100 Minecraft Diamond instructions across 25 length checkpoints (1 to 100 words, 4 per length) spanning navigation, collection, crafting, and survival categories, and measured whether the controller terminates at an appropriate time (judged by an API model), averaged over 5 seeds. The stop head is robust from $L{=}2$ through $L{=}50$, with no cliff at the training-distribution boundary ($L{=}11$), indicating generalization beyond training lengths. Degradation begins gradually above $L{=}50$, reaching $\sim$0.55 success at $L{=}100$; the only sharp drop is at $L{=}1$, where a single word carries insufficient semantic content to condition behavior.

\subsection{Training Dynamics of the Composite Loss}
\label{app:loss_dynamics}

\begin{figure}[t]
\centering
\includegraphics[width=0.85\textwidth]{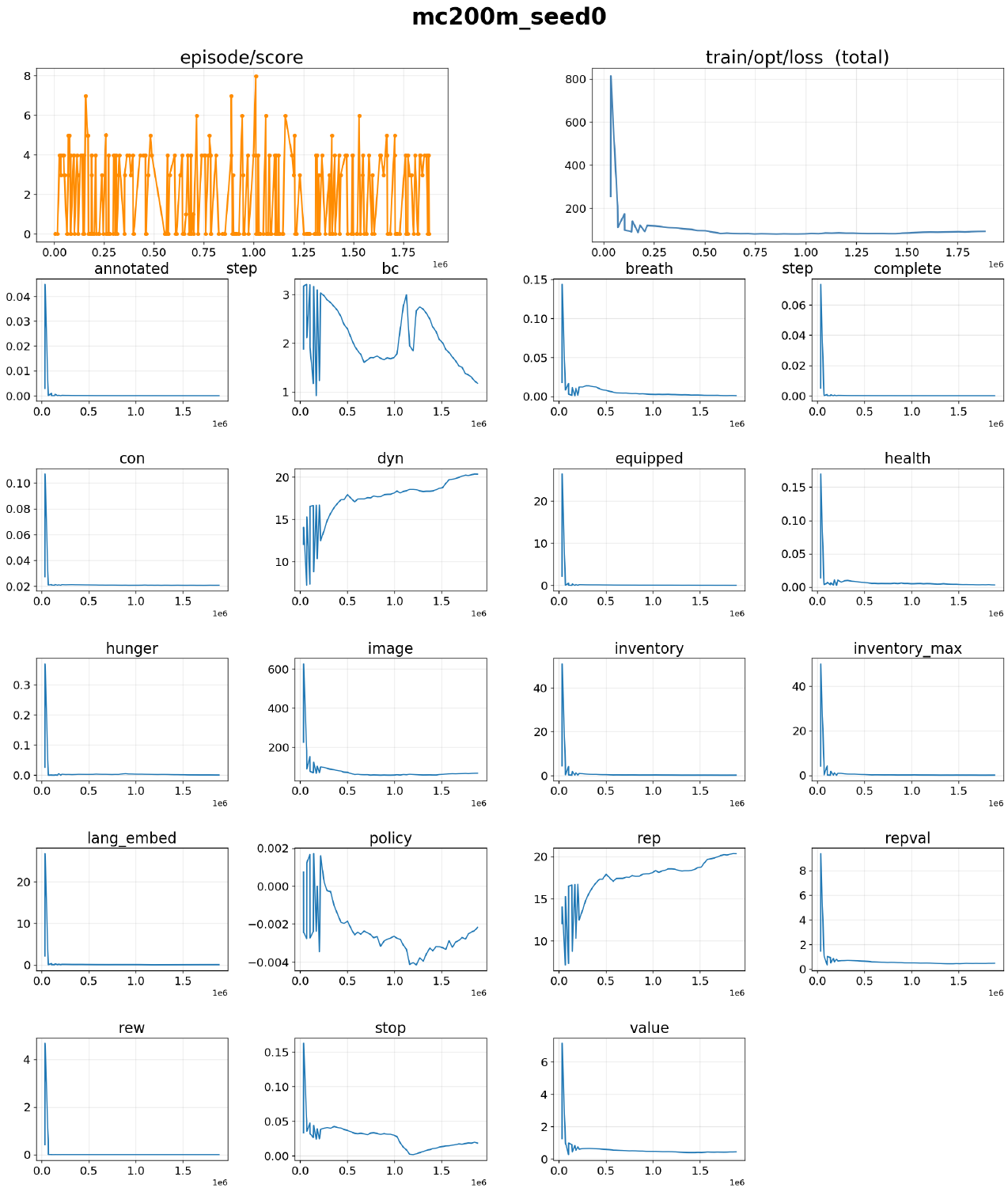}
\caption{Training curves of the individual loss components for a 200M-parameter controller on Minecraft Diamond (seed 0). The overall objective is a linear combination of the world-model losses (reconstruction, dynamics, and reward prediction), the actor and value objectives, the behavior-cloning (BC) loss on instruction-relabeled segments, and the stop-token loss.}
\label{fig:loss_dynamics}
\end{figure}

Figure~\ref{fig:loss_dynamics} shows how each sub-objective evolves over training. Early in training, the composite loss is dominated by the world-model reward term, as the controller has not yet learned to predict environment returns. As training progresses and instruction-relabeled segments accumulate in the replay buffer, the behavior-cloning loss becomes the dominant component. The dynamics loss exhibits an initial spike followed by a steady decrease, reflecting that the policy first encounters many unexpected observations during early exploration and subsequently learns the environment dynamics. Notably, the policy and behavior-cloning objectives continue to improve even as the expanding behavior distribution exposes the world model to new surprises, and we observed no single component collapsing or diverging. Training was stable across this run without loss re-weighting or scheduling beyond the fixed linear combination described in Section~\ref{sec:arch}.

\subsection{Prompt Details}
\label{app:prompts}

\begin{figure*}[ht]
\centering
\fbox{\includegraphics[width=0.85\textwidth]{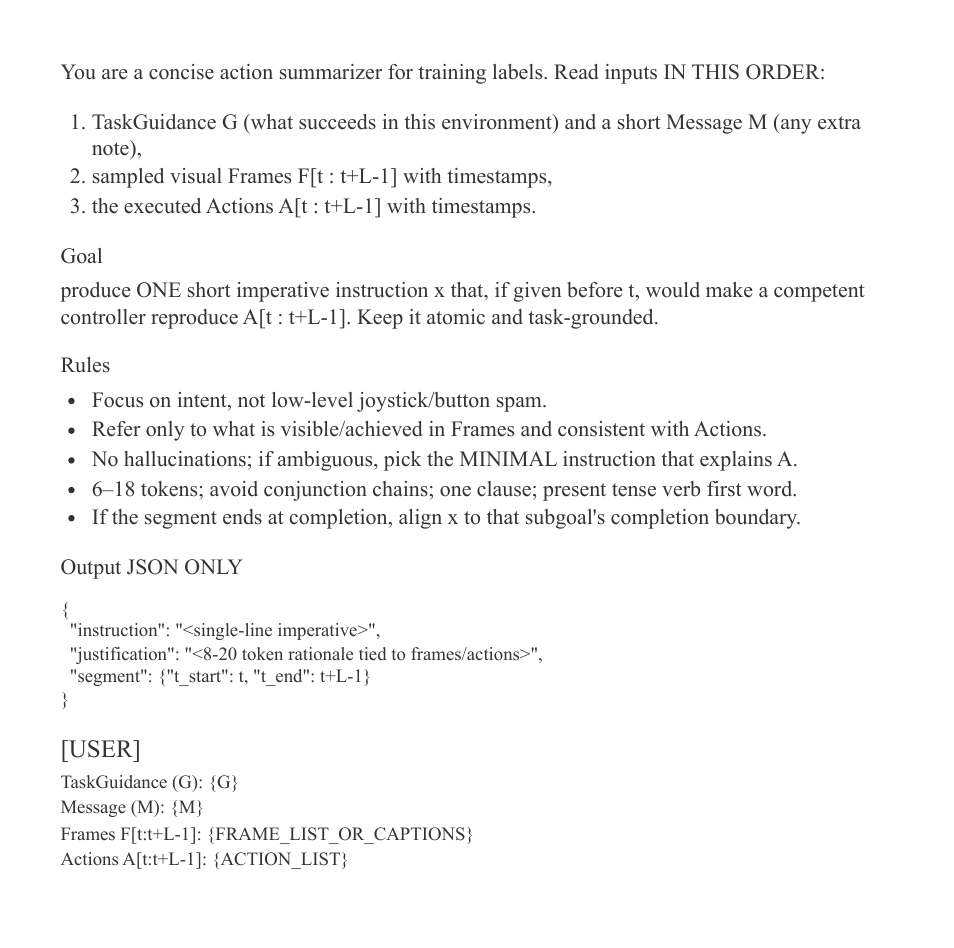}}
\caption{Training summarization prompt.}
\label{fig:sum_prompt}
\end{figure*}

\begin{figure*}[ht]
\centering
\fbox{\includegraphics[width=0.85\textwidth]{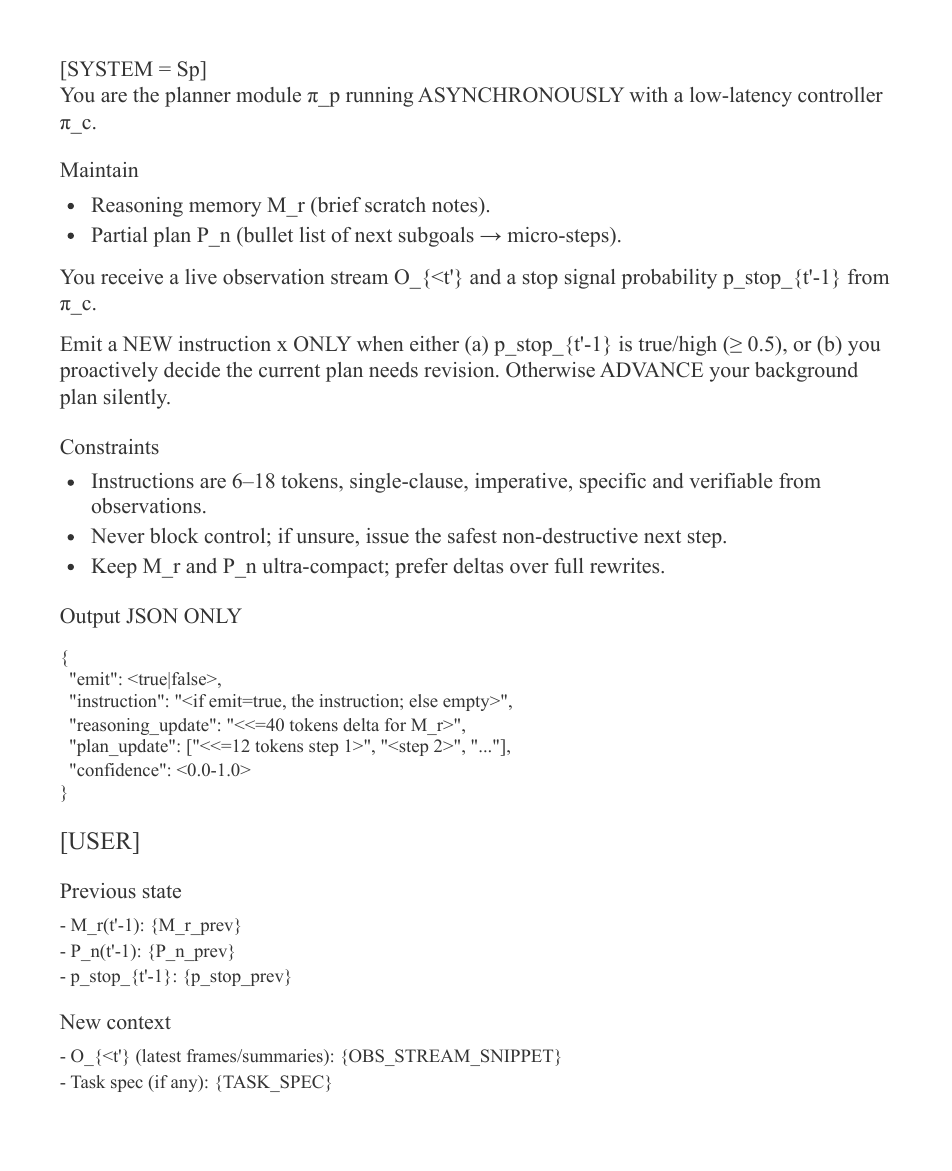}}
\caption{Single-agent prompt.}
\label{fig:single_prompt}
\end{figure*}

\begin{figure*}[ht]
\centering
\fbox{\includegraphics[width=0.85\textwidth]{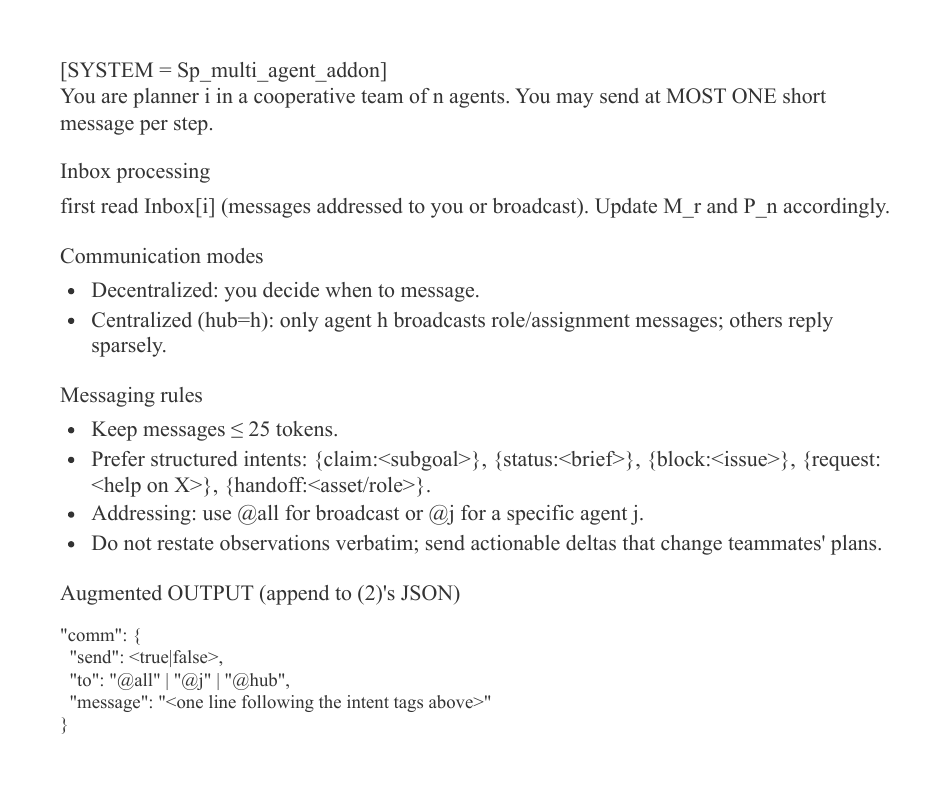}}
\caption{Multi-agent prompt (to be appended to single-agent prompt).}
\label{fig:multi_agent}
\end{figure*}

In this section, we list the prompts we used for training and inference including multi-agents.
The summarizer prompt for GPT-4o as shown in Fig.~\ref{fig:sum_prompt} is used for post-hoc training language annotation in Sec.~\ref{sec:generating-trajectories}. 
The inference prompt as shown in Fig.~\ref{fig:single_prompt} is used for online reasoning and issuing instructions as mentioned in Sec.~\ref{sec:online_inference}. 
The multi-agent prompt as shown in Fig.~\ref{fig:multi_agent} is used for prompting multi-agents communication. The full prompt is simply appending this to the single-agent inference prompt. Note that the math notations are all replaced by actual inputs or in-context examples.

\section{Tasks Details}
\label{app:task}
We list and describe the evaluation tasks in details here totaling 7 environments. Every environment can be discussed from its goal, the observation space, evaluation protocol, and the metric.
\paragraph{Arcade Learning Environment (ALE)}
\begin{itemize}
  \item\textbf{Goal:} Achieve high scores across Atari 2600 titles using standard ALE evaluation. 
  \item\textbf{Obs/Act:} Raw visual frames with common preprocessing (frame skip/stack); full primitive action set. 
  \item\textbf{Protocol:} Follow the evaluation practices popularized in recent Atari work, e.g., \citet{kaiser2019model,machado2018revisiting}, including fixed evaluation episodes and capped frames per episode; when using sticky actions or ALE “game flavours,” also follow the ALE protocol paper. 
  \item\textbf{Metric:} Game score per episode; report means (and variance) over seeds.
\end{itemize}

\paragraph{MineRL \texttt{ObtainDiamond}}
\begin{itemize}
    \item \textbf{Goal:} Obtain a diamond in survival Minecraft starting from scratch. 
    \item \textbf{Obs/Act:} $64{\times}64$ first-person RGB plus discrete inventory observations; actions combine continuous camera control with discrete navigation, mining/crafting/smelting interactions. 
    \item \textbf{Protocol:} Episode terminates on death, diamond obtained, or 18{,}000-frame (15 min) limit. Competition evaluation averages performance over 500 episodes on fixed but unseen seeds; strict train-from-scratch compliance. 
    \item \textbf{Metric:} Shaped milestone reward (e.g., logs, planks, pickaxes, iron, diamond), summed per episode; tie-break by fewest episodes to last milestone.
\end{itemize}

\paragraph{Crafter}
\begin{itemize}
\item \textbf{Goal:} Survive and progress in a procedurally generated 2D open world by unlocking semantically meaningful achievements (e.g., find water, craft tools, defeat enemies). 
\item \textbf{Obs/Act:} Local top-down pixel observations showing surroundings and inventory status; discrete actions for movement, interaction, crafting, sleeping, placing. 
\item \textbf{Protocol:} Two tracks: with extrinsic rewards and reward-free. Agents get a fixed budget of $1$M environment steps (commonly also reported at 5M in baselines). Success rates computed across the entire training run to emphasize sample efficiency. 
\item \textbf{Metric:} \emph{Crafter score} is the geometric mean of the 22 achievement success rates, emphasizing breadth/depth of capabilities.
\end{itemize}

\paragraph{DeepMind Lab: \texttt{explore\_goal\_locations}}
\begin{itemize}
\item \textbf{Goal:} First-person 3D navigation in maze-like levels to discover and reach goal locations under partial observability. 
\item \textbf{Obs/Act:} RGB first-person frames; continuous look and movement controls. 
\item \textbf{Protocol:} Standard DM-Lab evaluation with fixed episode caps; commonly evaluated on held-out maps (unseen layouts), sometimes with environment-provided debug info \emph{only} for visualization/analysis (not agent inputs). 
\item \textbf{Metric:} Episode return (goals found/reached), success rate, and path efficiency; report means over seeds/maps.
\end{itemize}

\paragraph{Overcooked-AI}
\begin{itemize}
\item \textbf{Goal:} Two-player cooperative cooking (deliver soups quickly) with strong \emph{zero-shot coordination} (ZSC) to unseen partners and layouts. 
\item \textbf{Obs/Act:} Gridworld state or egocentric features; discrete actions (move, interact, pick/place). 
\item \textbf{Protocol:} Evaluate on canonical layouts (\textit{Cramped Room}, \textit{Asymmetric Advantages}, \textit{Coordination Ring}, etc.) and out-of-distribution layouts; cross-play with behavior-cloned human models and held-out agents; human studies where applicable. 
\item \textbf{Metric:} Team return (deliveries/time), success rate, and ZSC scores (cross-play averages across unseen partners/layouts).
\end{itemize}

\paragraph{Pico Park (Co-op Puzzles)}
\begin{itemize}
\item \textbf{Goal:} Complete short, cooperative puzzle-platforming levels that demand synchronized actions (e.g., stacking, shared switches, tethered movement). 
\item \textbf{Obs/Act:} Platformer state with simple discrete controls for 2–8 players (levels scale with player count). 
\item \textbf{Protocol:} Curate a fixed subset of multiplayer levels (e.g., 48 classic levels) and evaluate multiple seeds/player configurations; require that all agents reach the goal to clear a level. 
\item \textbf{Metric:} Level completion rate and median completion time across the suite; optionally, coordination error counts (drops, desyncs).
\end{itemize}

\paragraph{MindCraft (Minecraft Multi-Agent Collaboration)}
\begin{itemize}
\item \textbf{Goal:} Multi-agent embodied collaboration on \emph{Cooking}, \emph{Crafting}, and \emph{Construction} tasks via language-enabled coordination. 
\item \textbf{Obs/Act:} First-person Minecraft control with inventories; agents exchange natural-language messages; tasks provide recipes/blueprints and split resources/knowledge across teammates. 
\item \textbf{Protocol:} Procedurally generated tasks per category; for construction, initialize agents with blueprints and disjoint materials/skills; for cooking/crafting, vary recipe complexity and information asymmetry (``Hell's Kitchen'' variants). 
\item \textbf{Metric:} Average success rate for Cooking \& Crafting; Construction uses an edit-distance alignment between the built structure and the target blueprint; overall score averages category scores.
\end{itemize}

\section{Algorithm}
\label{app:algorithm}

\begin{center}
\begin{minipage}[t]{0.48\textwidth}
\begin{algorithm}[H]
\caption{Online planning at timestep $t$}
\label{alg:online}
\begin{algorithmic}[1]
\footnotesize
\State \textbf{Input:} environment $P$, controller $\pi_c$, VLM planner $\pi_p$, \texttt{Stop} $\in \{0,1\}$, \texttt{Inbox} $\in \mathcal{X}$, \texttt{Plan} $\in \mathcal{X}$, latest observation $o_t$.
\medskip
\State \hspace{0em}\textbf{thread Controller():}
\State \hspace{1em}\textbf{while} running \textbf{do}
\State \hspace{2em} $\left(a_t, s_t\right) \gets \pi_c.\mathrm{step}\left(o_t, x_t\right)$
\State \hspace{2em} $P.\mathrm{step}(a_t)$
\State \hspace{2em} \textbf{if} $s_t$ \textbf{then} \texttt{Stop} $\gets$ 1; $x_{t+1} \gets \varnothing$
\State \hspace{2em} \textbf{if} \texttt{Inbox} \textbf{then} $\mathbf{x}_{t+1} \gets$ \texttt{Inbox}
\medskip
\State \hspace{0em}\textbf{thread VLM():}
\State \hspace{1em}\textbf{while} running \textbf{do}
\State \hspace{2em} \textbf{if} \texttt{Stop} \textbf{then}
\State \hspace{3em} \texttt{Inbox} $\gets \pi_p.\mathrm{emit}(o_t,\texttt{Plan})$
\State \hspace{3em} \texttt{Stop} $\gets 0$; \texttt{Plan} $\gets \varnothing$
\State \hspace{2em} \textbf{else}
\State \hspace{3em} \texttt{Plan} $\gets \pi_p.\mathrm{advance}(o_t,\texttt{Plan})$
\end{algorithmic}
\end{algorithm}
\end{minipage}%
\hfill
\begin{minipage}[t]{0.48\textwidth}
\begin{algorithm}[H]
\caption{Offline planning at timestep $t$}
\label{alg:offline}
\begin{algorithmic}[1]
\footnotesize
\State \textbf{Input:} environment $P$, controller $\pi_c$, VLM planner $\pi_p$,  \texttt{Stop} $\in \{0,1\}$, latest observation $o_t$, plan state $Pn_t$.
\medskip
\State \textbf{while} running \textbf{do}
\State  \hspace{1em}{$\triangleright$\textit{Add to the current plan, potentially emitting an instruction.}}
\State \hspace{1em} $(x_t, Pn_{t+1}) \gets \pi_p\mathrm{.step}(o_t, Pn_t, \texttt{Stop})$
\State \hspace{1em} \textbf{if} $x_t$ \textbf{then} \texttt{Stop} $\gets 0$
\medskip
\State \hspace{1em} $\triangleright$\textit{Execute the existing instruction.}
\State \hspace{1em} $(a_t, s_t) \gets \pi_c.\mathrm{step}(o_t, x_t)$
\State \hspace{1em} $P.\mathrm{step}(a_t)$
\State \hspace{1em} \textbf{if} $s_t$ \textbf{then} \texttt{Stop} $\gets$ 1
\end{algorithmic}
\end{algorithm}
\end{minipage}
\end{center}

\paragraph{Shared-state semantics.}
Inbox, Stop, and Plan are synchronized through step-indexed buffers rather than arbitrary cross-thread writes. The controller writes a completion event when $p_{\text{stop}}$ crosses its threshold. If this happens while the planner is mid-generation, the planner finishes its current \textsc{advance} call using the latest available observation context and stores the resulting draft plan in Plan. At the planner's next scheduling point, the completion event triggers a short \textsc{emit} call that converts the draft into the next executable instruction and writes it atomically to Inbox; the controller takes in the latest instruction at the next control step. Observation staleness is therefore bounded by the duration of the planner's current background reasoning call plus one control-step boundary, and the controller continues executing its previous instruction throughout, so staleness affects only the next high-level instruction and never low-level control. In terms of token cost, \textsc{advance} carries most of the reasoning tokens, as it updates the planner's memory and partial plan while the controller acts, whereas \textsc{emit} only decodes a short instruction from the existing draft. This asymmetry is why asynchronous planning reduces instruction-boundary latency relative to blocking inference (quantified in Table~\ref{tab:async}).

\paragraph{Multi-agent chatroom synchronization.}
Each agent runs its own asynchronous planner thread and controller thread. Planner outputs, including chat messages and instructions, are never injected into the environment at arbitrary wall-clock times. Instead, they are written to a per-agent mailbox and consumed at the next synchronized communication/control boundary, at which point each agent reads the latest incoming messages and updates its planner context. Communication thus proceeds in discrete rounds aligned with control steps while planner reasoning proceeds asynchronously in the background.

\paragraph{Planner observability.}
In all environments, the planner receives only the observations available to the agent itself. No privileged simulator state, full-state access, or hidden information is provided, ensuring the framework applies to settings where privileged information is unavailable.

\section{Instruction-following evaluation protocol.}
\label{app:human}

Instruction-following accuracy was judged by one of the authors on 200 held-out planner-generated instructions per domain. The evaluator was shown the instruction, the corresponding observation/trajectory segment, and a real-time video of the controller's execution, and judged whether the behavior completed the intended instruction. As a sanity check, we repeated the same annotation task with an API model (GPT-5.5), which agreed with the human annotator on 93.2\% of cases across tasks. We emphasize that single-annotator judgment combined with model agreement serves as a sanity check rather than a substitute for independent multi-annotator human evaluation, and we present these accuracies with that caveat.

\subsection{Additional Training and Communication Details}
\label{app:training_details}
When collecting a rollout of length $T$, different timesteps $t$ may be associated with different instruction embeddings $e_t$ and completion indicators $\mathrm{complete}_t$, whereas $\mathrm{done}_t$ is true if and only if $t = T$. To sample annotation segments, a segment length $L$ is first randomized within an integer interval, then a start index is drawn as $t \sim \mathrm{Uniform}\{1,\dots,|\mathcal{D}|{-}L\}$; sampled segments are constrained to be non-overlapping.

Formally, for multi-agent communication with $n$ planner agents, at each VLM inference step any planner $i$ may initiate at most one message $m_{ij}$ to planner $j$ in the decentralized mode. In the centralized mode, the sender is a fixed hub agent $h$ that sends messages $m_{hi}$ to all $i \in [n]\setminus h$.
\end{document}